%% file: main.tex
\documentclass[10pt,twocolumn,letterpaper]{article}

\usepackage{cvpr}              % To produce the CAMERA-READY version
\usepackage{booktabs}
\usepackage{multirow}
\usepackage{graphicx}
\usepackage{algorithm}
\usepackage{algpseudocode}
\usepackage{amsmath}
\usepackage{stfloats}

\usepackage{times}
\usepackage{epsfig}
\usepackage{graphicx}
\usepackage{amsmath}
\usepackage{amssymb}
\usepackage{multirow}
\usepackage{algorithm}
\usepackage{algpseudocode}
\usepackage{booktabs}
\usepackage{color}
\usepackage{xcolor}
\usepackage{cuted}
\usepackage{capt-of}
\usepackage[normalem]{ulem}
\usepackage{longtable}
\usepackage{multirow}
\usepackage{cite}
\usepackage{tabularx}


\definecolor{cvprblue}{rgb}{0.21,0.49,0.74}
\usepackage[pagebackref,breaklinks,colorlinks,citecolor=cvprblue]{hyperref}

\def\paperID{20} % *** Enter the Paper ID here
\def\confName{CVPR}
\def\confYear{2024}

\title{Practical Region-level Attack against Segment Anything Models}

\author{Yifan Shen\thanks{Equal Contribution} \quad Zhengyuan Li\footnotemark[1] \quad Gang Wang \vspace{0.1em} \\ 
    University of Illinois Urbana-Champaign \vspace{0.1em}\\
    {{\tt \{yifan26, zli138, gangw\}@illinois.edu}}
}
\begin{document}
\maketitle

\begin{strip}
\centering
\vspace{-0.1in}
\includegraphics[width=0.75\textwidth]{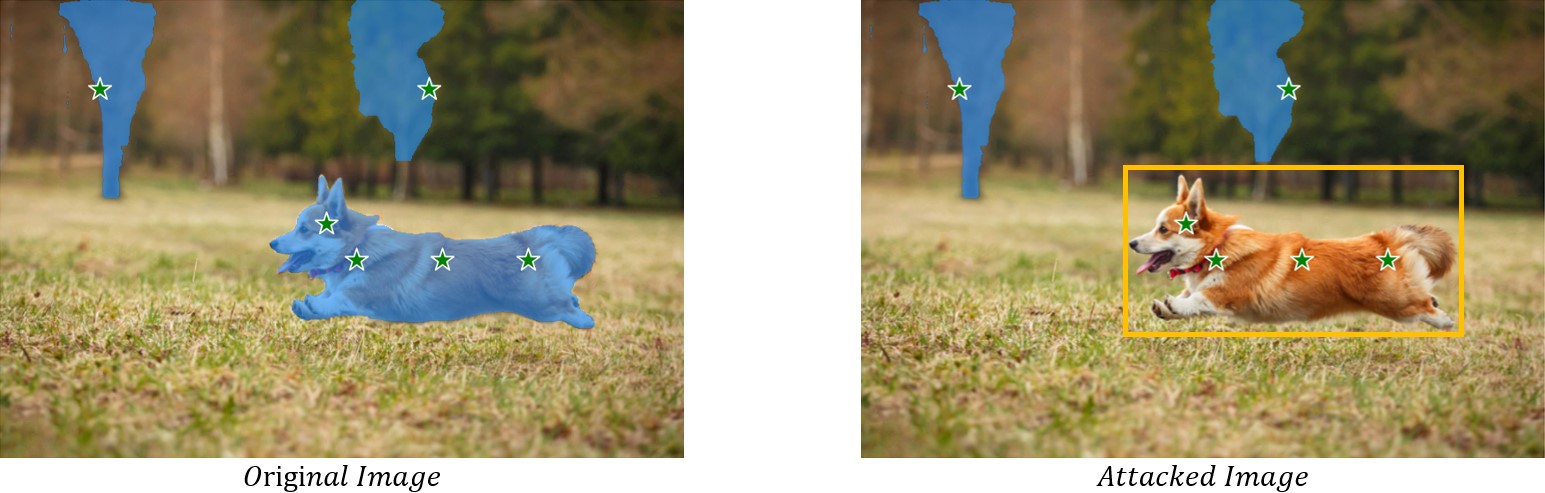}
\captionof{figure}{\textbf{Region-level Attack on a Segment Anything Model (SAM).} The left image shows the original {\em clean} image---objects are well segmented when a user clicks on the object region (user clicks are denoted by green stars). The right image shows the {\em attacked} image---the corgi in the yellow box (attack-target region) can no longer be identified by SAM no matter where the user clicks within the box. Note that, the regions outside of the yellow box in the image are not affected by the attack.}
\vspace{-0.1in}
\label{fig:teaser}
\end{strip}

\input{sec/0_abstract}    
\input{sec/1_introduction}

\input{sec/2_relatedwork}
\input{sec/3_preliminary}
\input{sec/3_method}

\input{sec/4_experiments}
\input{sec/5_transfer}
\input{sec/6_discussion}

% \input{sec/7_conclusion}
\noindent{\footnotesize\textbf{Acknowledgments.} This work was supported in part by NSF grants 2055233 and 2229876. Any opinions, findings, and conclusions expressed in this material are those of the authors and do not necessarily reflect the views of the funding agencies.}
% \newpage

{
    \small
    \bibliographystyle{ieeenat_fullname}
    \bibliography{main}
}


\end{document}

%% file: sec/0_abstract.tex
\begin{abstract}
Segment Anything Models (SAM) have made significant advancements in image segmentation, allowing users to segment target portions of an image with a single click (i.e., user prompt). Given its broad applications, the robustness of SAM against adversarial attacks is a critical concern. While recent works have explored adversarial attacks against a pre-defined prompt/click, their threat model is not yet realistic: (1) they often assume the user-click position is known to the attacker (point-based attack), and (2) they often operate under a white-box setting with limited transferability. In this paper, we propose a more practical region-level attack where attackers do not need to know the precise user prompt. The attack remains effective as the user clicks on any point on the target object in the image, hiding the object from SAM. Also, by adapting a spectrum transformation method, we make the attack more transferable under a black-box setting. Both control experiments and testing against real-world SAM services confirm its effectiveness. Code is released at \href{https://github.com/ShenYifanS/S-RA_T-RA}{https://github.com/ShenYifanS/S-RA\_T-RA}.
\end{abstract}

% Recently, Segment Anything
% Model (SAM), has become a famous promptable segmentation model that enables users to segment portions of an image with a single click. The vulnerability of the SAM model to adversarial attacks has sparked significant concerns regarding its robustness. 
% \gang{Explain why SAM stand for when it first appears}\Zhengyuan{revised}
% However, existing threat models assume that the precise position of the point prompt is known to the attacker, which is not likely in practice.
% \gang{Need to point out exactly why the point-based attack is not practical, we should say what assumptions they have that make the attack less practical.}\Zhengyuan{revised}
% In practice, attackers might seek to render SAM oblivious to entire objects. To address this, we introduce the region-level Attack, a novel threat model that elevates the attack granularity from a single point to an entire region. The attack model seeks a universal attack for all points within a user-defined region $S$. In addition, we propose T-RA, a method that generates transferrable adversarial samples. Through extensive experiments, we demonstrate that: (1) Region-level attacks are feasible in both white-box and black-box settings. (2) Our method generates adversarial samples against multiple SAM variants.

%% file: sec/1_introduction.tex
\section{Introduction}
\label{sec:intro}

Segment Anything Models (SAM) leverage foundational models for {\em promtable} image segmentation~\cite{kirillov2023segment}, which has shown outstanding performance. SAM can delineate objects of interest into masks based on user prompts (e.g., point of user clicks). As SAM is used for mission-critical applications such as healthcare image analysis~\cite{ma2024segment, mazurowski2023segment} and scene understanding for autonomous driving~\cite{zhao2023enhancing}, the robustness of SAM (against adversarial attacks) raises concerns. For example, adversaries may manipulate the images/videos taken by autonomous vehicles to hide key objects (e.g., traffic signs, vehicles) from the image segmentation module, posting threats to driving safety.

While recent work has explored adversarial attacks against SAM~\cite{zhang2023attack,zheng2023black}, their threat models are not yet realistic. For instance, Attack-SAM~\cite{zhang2023attack} proposes a white-box attack to hide an object in the image from being segmented by SAM. However, Attack-SAM assumes the precise {\em point location} where the user clicks and the {\em model parameters} are both known to the attacker. Sheng \etal~\cite{zheng2023black} investigated another attack (TAA) under a different threat model. Their goal is to mislead SAM to output a mask of attacker-specified shape---instead of hiding the object from SAM (which is our main focus). Their results also show that this is a challenging problem: while TAA has some transferability, the attack effect is majorly weakened after transferring. 

In this paper, we introduce a region-level attack to explore a more practical threat model (see \cref{fig:teaser}). The attacker's goal is to conceal the object within an attacker-specified region from SAM's segmentation. In this case, the attacker does not need to know the precise point of the click of the user---no matter which point in the region is clicked by the user, the object cannot be accurately segmented by SAM. In addition, we investigate to improve the transferability of the attack such that it can operate under a {\em black-box} setting. 
As shown in \cref{fig:teaser}, after applying the adversarial perturbation to the image, clicking on any point in the yellow box (attacker-specified region) will no longer separate the corgi from the rest of the image.

Under this threat model, we first develop a Sampling-based Region Attack (S-RA), a basic method for region-level adversarial attacks, and then improve its transferability with a Transferable Region Attack (T-RA). Our design is based on two key intuitions. First, sparsely sampled points in the region can constitute a surrogate target of all pixels in the region. Second, even when the involved region goes beyond a single point, adding perturbations in the frequency domain when attacking the surrogate model can improve the transferrability~\cite{long2022frequency}.
Therefore, our method first applies spectrum transformation to the image in order to simulate the spectrum saliency map~\cite{long2022frequency} of the victim model. Then it estimates the optimization target with evenly sampled points in the region and conducts the optimization with a PGD attack~\cite{madry2017towards} to generate adversarial noises.

We evaluated the proposed attacks on multiple SAM variants including ViT-B, ViT-H and ViT-L~\cite{kirillov2023segment} and demonstrated the effectiveness of the attacks under both white-box and black-box settings. We extensively evaluated the attack transferability to a variety of SAM architectures including EfficientSAM (S and Ti)~\cite{xiong2023efficientsam}, Fast-SAM (S and X)\cite{zhao2023fast}, MobileSAM~\cite{mobile_sam}, and HQ-SAM (B, L, and H)~\cite{sam_hq}. 
We also confirm the effectiveness of the attack (optimized with a local ViT-B) against a real-world SAM service. Our result highlights the realism of the risk and calls for new defense methods to improve the robustness of SAM. 

Our contributions are summarized as follows.

\begin{itemize}
    \item We present a region-level attack against SAM, a more practical threat model where attackers do not need to know the precise user prompt. 
    
    \item We designed novel attack methods, Sampling-based Region Attack (S-RA) and Transferable Region Attack (T-RA), that undermine SAM's segmentation ability under both white-box and black-box settings. 
    \item Extensive experiments demonstrate that S-RA and T-RA can successfully attack the original SAM and its variants.
\end{itemize}

%% file: sec/2_relatedwork.tex
\section{Related Works}\label{sec:rela}
\subsection{Adversarial Attacks}
Deep neural networks are known to be susceptible to adversarial examples, which are samples that appear indistinguishable from genuine ones to the human eye but can mislead models into producing incorrect outputs~\cite{szegedy2013intriguing, biggio2013security,zhuang2023pilot}.

Attacks are manifested in two settings: white-box and black-box. 
In the white-box setting, attackers can access all model knowledge, including architecture, parameters, and gradients. This setting is often used to assess model robustness rather than actual attacks~\cite{goodfellow2014explaining, carlini2017towards, madry2017towards}. White-box attacks, such as Fast Gradient Sign Method (FGSM)~\cite{goodfellow2014explaining} and projected gradient descent (PGD)~\cite{madry2017towards}, allow full visibility into the target model to generate adversarial examples. 
In contrast, black-box attacks operate under limited knowledge~\cite{guo2020backpropagating, wu2020skip, xie2019improving, zhang2022investigating}. Notable techniques include updating gradients with momentum (MI-FGSM)~\cite{dong2018boosting}, smoothing gradients with a kernel (TI-FGSM)~\cite{dong2019evading}, and resizing adversarial examples for input diversity (DI-FGSM) ~\cite{xie2019improving}. These methods are grounded in the principle of the transferability of adversarial examples exploiting vulnerabilities inherent across multiple models without specific insights into the target model's internals~\cite{zhao2021success, zhang2022investigating}.

% More recently, it has been shown that combining the above techniques constitutes simple yet strong approaches for transferable targeted attacks ~\cite{zhao2021success, zhang2022investigating}. 
% Adversarial attacks can also be categorized by the attack goal. For image recognition tasks, ``untargeted'' attacks aim for any deviation from the ground truth. In contrast, ``targeted'' attacks, with a stricter goal, seek to mislead the model into producing a specific, predetermined output. 

% Researchers have explored methods to compute adversarial perturbations with minimal $L_2$ and $L_\infty$ norms~\cite{cisse2017houdini,laidlaw2020perceptual}. As evinced by the recent research in attack strategies~\cite{zhuang2023pilot, wang2023global, lei2023sgu, xie2017adversarial}, achieving robustness against adversarial attacks remains an open challenge.

% \cite{ozbulak2019impact,wang2023global,xie2017adversarial}

% \cite{lei2023sgu}

% Meanwhile, , could be repurposed for segmentation. By leading models to erroneously label significant portions of images, this study, along with others (\cite{cisse2017houdini, xie2017adversarial}), illuminated the pervasive vulnerabilities present in sophisticated deep models. The intricate adversarial challenges posed to models like SAM~\cite{zhang2023attack, rony2023proximal}, further underscore the evolving complexities in this domain. 

\subsection{Segment Anything Model and Variants}
In the domain of image segmentation, major advancement has been made by the ``Segment Anything Models'' (SAM)~\cite{kirillov2023segment}. SAM uses foundational models and capitalizes on the principles established by prior works, underscoring the importance of multi-scale features and iterative refinement for segmentation~\cite{zhao2017pyramid, ronneberger2015u, badrinarayanan2017segnet}. SAM's versatility is further explored through its application across diverse contexts, for example, medical image segmentation~\cite{ma2023segment}, detection of camouflaged entities~\cite{tang2023can}, and semantic communication challenges~\cite{tariq2023segment}.
% demonstrating its efficacy in not only conventional settings but also in more complex scenarios. 
% The exploration extends to applying SAM's segmentation capabilities for 
SAM has been applied to both 2D and 3D environments, highlighting its potential in semantic labeling~\cite{chen2023semantic}, object tracking~\cite{yang2023track, zhang2023uvosam}, and 3D object segmentation~\cite{shen2023anything3d, chen2023voxenext}.

In recent variants of SAM, SEEM~\cite{zou2024segment} allows users to segment images using various ``prompts'', including points, markers, boxes, scribbles, text, and audio. HQ-SAM~\cite{sam_hq} enhances the ability to accurately segment any object. Semantic-SAM~\cite{li2023semantic} emerges as a universal image segmentation model that enables segmentation and recognition at any desired granularity. For improved efficiency, MobileSAM~\cite{mobile_sam} introduces object-aware prompt sampling, replacing the grid-search prompt sampling in the original SAM, to expedite the segmentation process. EfficientSAM~\cite{xiong2023efficientsam} leverages Masked Image Pretraining to improve segmentation efficiency. Fast Segment Anything~\cite{zhao2023fast} speeds up the original SAM model by 50x. 
% These models collectively advance image segmentation technology, offering diverse solutions tailored to varying application requirements.

Prior works have assessed the robustness of {\em traditional} segmentation models against adversarial attacks~\cite{fischer2017adversarial,kang2020adversarial,ozbulak2019impact,wang2023global,xie2017adversarial}. More recent works have explored the problem in the context of SAM~\cite{zhang2023attack, rony2023proximal}, using {\em imperceivable} adversarial perturbations.  
Yu \etal~\cite{qiao2023robustness} introduce an attack that produces {\em visible} image corruptions such as style changes, occlusions, and local patch attacks.
However, as discussed in \cref{sec:intro}, the existing SAM attacks' threat model is not yet realistic by assuming knowing precise user prompt (under a white-box setting), and we aim to improve the realism of the attack with a region-based attack (black-box setting).

% In the context of our research, the robustness of SAM has been extensively examined across various scenarios. Attack-SAM~\cite{zhang2023attack}'s threat model is limited to single points. Sheng \etal ~\cite{zheng2023black} investigates targeted attacks based on user-defined target images. Beyond adversarial attacks, Yu \etal~\cite{qiao2023robustness} primarily concentrates on visible image corruptions such as style changes, occlusions, and local patch attacks. Diverging from prior investigations, our approach introduces an untargeted and realistic threat model that extends from individual points to entire regions. The adversarial attacks generated by our method are transferable and imperceptible.

% \Zhengyuan{revised}
% \gang{vulnerable or not? 
% Briefly remind people how we are different/novel: 
% region attack, more realistic
% transferable over black-box, more realistic. 
% }

% \gang{Where do we talk about point-based SAM Attack?}\Zhengyuan{sec. 4.1}

% \gang{We need to discuss our threat model somewhere in the paper: 
% Attacker's goal, assumptions, knowledge, etc. White-box vs. black-box, etc. 
% }\Zhengyuan{sec 4.1}

%% file: sec/3_preliminary.tex
\section{Preliminary}
\label{sec:pre}

\paragraph{Segment Anything Model (SAM).}
SAM introduces a novel, promptable segmentation framework to generate precise masks for a given image and a prompt. 
While the original SAM~\cite{kirillov2023segment} supports points, text, and boxes as prompts, in this work, we primarily focus on the point prompt scenario (similar to \cite{zhang2023attack,zheng2023black}), leaving other types of prompts for future exploration.
% \gang{is the prompt a ``text'' or a point on the image?}\Zhengyuan{revised}
At its core, SAM comprises three key components: 
an image encoder, a prompt encoder, and a mask decoder. 
The image encoder leverages a Vision Transformer (ViT) architecture, pretrained using the Masked Autoencoder (MAE)~\cite{he2022masked}, to extract feature representations from input images. 
The prompt encoder employs positional embeddings to encode prompts. 
The mask decoder synthesizes the outputs from both encoders to predict segmentation masks, thus determining the segmented object based on the synergy between the image and the prompt. The mask prediction process in SAM is defined as follows:
\begin{align}
    y = \textit{SAM}(p, x; \theta)
\end{align}
where \(p\) and \(x\) denote the input prompt and image, respectively, and \(\theta\) symbolizes the model's parameters. Given an image \(x \in \mathbb{R}^{H \times W}\), the output \(y\) mirrors the input image's dimensions, with \(H\), and \(W\) representing the height, and width, respectively. The pixel coordinates within image \(x\) are denoted by \(i\) and \(j\). The mask region is delineated by the predicted values \(y_{ij}\) where values exceeding a defined threshold (e.g., 0) are classified as part of the segmented object. 
% \gang{y's value is 1 or 0?}
% \Zhengyuan{real value}
% \Zhengyuan{Deleted: This study predominantly explores the efficacy of SAM with basic point inputs. }
% \gang{why? explain the reason}
During inference, the final binary mask is obtained as follows, where \(\textit{sign}\) denotes the sign function.
\begin{align}
    M_{pred} = \textit{sign}(y)
\end{align}

\paragraph{Projected Gradient Descent Attack.}
Projected gradient descent (PGD) attack~\cite{madry2017towards} is a popular adversarial attack method. It utilizes the first-order gradient and iteratively finds the solution to the optimization problem within the allowed perturbation set. The algorithm can be concisely represented as follows:
\begin{align}
    x^{(t+1)} = \textit{Clip}_{x,\epsilon}\left(x^{(t)} + \alpha \cdot \textit{sign}(\nabla_x L(\theta, x^{(t)}, y))\right)
\end{align}
where \( x^{(t)} \) is the adversarial example at iteration \( t \), \( \alpha \) is the step size, \( L \) is the loss function defined by the specific task, \( \theta \) represents the model parameters, and \( y \) is the ground-truth label. 
% \gang{True label or target label?}\Zhengyuan{true}
The function \( \textit{Clip}_{x,\epsilon} \) ensures that the perturbed image remains within an \( \epsilon \)-neighborhood of the original image, which means $ \lVert \delta \rVert_{\infty} < \epsilon$. In other words, $\epsilon$ controls the magnitude of the adversarial perturbation. The method is untargeted (i.e., the adversarial example aims to obtain any other labels that are not the ground-truth label $y$). It serves as a fundamental building block of our attack method.

%% file: sec/3_method.tex
\section{Method}\label{sec:method}

% \subsection{Threat Model and Attack Overview}
% \gang{Sample text below: TODO: revise to put some real text.}\Zhengyuan{revised}
In this section, we introduce the threat model and our detailed attack method. We start with the basic point-level attack setting and then build the idea of region-level attacks on top of it, and then use spectrum transformation to further improve the transferability. 

\subsection{Threat Movel and Attack Overview}

\paragraph{Point-level Attack.}  Point-level attack~\cite{zhang2023attack} was previously proposed to attack SAM, assuming the attacker knows the user prompt (i.e., click point). The goal of the attack is to conceal a target object, which is formulated as minimizing the prediction values of the SAM-generated mask \(y\). Formally, given an image $x$ and a set of permissible adversarial perturbations, \(S\), constrained by predefined attack strength parameters, the goal is to compute the adversarial perturbation $\delta$ that obliterates the mask when the model is prompted with a specific point ($p$). The loss function is defined as:
\begin{align}\label{formula:loss}
    L(x, p) &=  \left\|\textit{Clip}\left(\textit{SAM}(p, x+\delta; \theta), \right.\right. \nonumber \\
    &\quad \left.\left. \textit{min} = Neg_{th}\right) - Neg_{th}\right\|^2,
\end{align}
where $\theta$ is the parameters of the target model, and \(Neg_{th}\) is a negative hyperparameter threshold used by SAM to segment an object (see \cref{sec:pre}). Following~\cite{zhang2023attack}, we set 
\(Neg_{th}\)to -10, as the non-mask regions often have values around this threshold. The point-level threat model is the following optimization problem:
\begin{align}
    \delta^* = \arg\min_{\delta \in S} L(x,p)
\end{align}
where $\delta^*$ represents the optimal value of $\delta$, which is the perturbation that achieves the best adversarial effect.
The application of the clip function is strategic, preventing the predictive values from becoming excessively negative, which could inadvertently impede the optimization process. This ensures that predicted values, \(\textit{SAM}(p, x + \delta)\), are coerced towards being less than or equal to \(Neg_{th}\), with clipping applied to maintain values above this threshold.

\paragraph{Region-level Attack.} Region-level attack allows the user to specify a region $R$ where SAM should fail to segment the objects regardless of the user prompts. Formally, 
\begin{align}\label{formula:region}
    \delta^* = \arg\min_{\delta \in S} \mathbb{E}_{p\sim \text{uniform}(R)}[L(x,p)]
\end{align}
which means for a point uniformly sampled from $R$, we minimize the expectation of SAM's segmentation mask's area. In this work, we assume that $R$ is a rectangle that covers the object specified by the attacker.

\subsection{Sampling-based Region Attack (S-RA)}
First, we discuss a {\em white-box} attack. 
Directly optimizing ~\cref{formula:region} is computationally intensive due to the large number of pixels in an image. Alternatively, we sample points in the region and create the substitute loss function. We uniformly select points by partitioning the $R$ into a grid where \(m\) points are chosen along the horizontal axis and \(n\) points along the vertical axis, resulting in a total of \(m \times n\) points. The loss function for the point set is
\begin{align}\label{formula:loss_set}
    L_{SRA}(x, P) &=  \frac{1}{m \times n}\sum_{p\in P} L(x,p)
\end{align}
Compared with random sampling used in previous work~\cite{zheng2023black}, this structured selection process ensures comprehensive coverage of the targeted region. Each point within this grid is subsequently targeted with the point attack strategy. 

% This structured selection process ensures a comprehensive coverage of the targeted region. Each point within this grid is subsequently targeted with the point attack strategy. The cumulative impact of the Region Attack is quantitatively evaluated by measuring the overall deviation in the segmentation mask predictions for the \(N\) selected points, compared to the mask of the original, unperturbed image. This is mathematically represented as:
% \begin{align}
%     \Delta M = \sum_{i=1}^{N} \left|\textit{SAM}(p_i, x+\delta_i; \theta) - \textit{SAM}(p_i, x; \theta)\right|,
% \end{align}
% where \(p_i\) is the \(i\)-th point in the grid, \(x\) denotes the original image, \(x + \delta_i\) indicates the image perturbed at point \(p_i\), and \(\theta\) represents the model parameters. 

In the later experiments (\cref{sec:exp}), we will evaluate the attack effectiveness by {\em randomly} selecting a point within the region $R$ as a prompt to examine the segmentation result.
Note that, the newly selected point during testing time is not necessarily (unlikely) among these sampled points used for attack optimization, due to the sparsity of sampling.

% \subsection{Black-box: T-RA}
% in transferring adversarial examples from one model to another, for example, when evaluating the transferability from a small Vision Transformer (ViT-B) to a large one (ViT-H). This limitation underscores the necessity for a more robust attack methodology capable of 

\subsection{Transferable Region Attack (T-RA)}
Under the {\em black-box} setting, attackers need to compute adversarial perturbations based on a local substitute model and then apply the perturbation to attack a different target model. 
% A paramount challenge is to bridge the gap between the substitute model and the target model. 
The above sample-based region attack (S-RA) has shown limited transferability (see \cref{sec:exp}), which motivates us to improve it for black-box attacks. More specifically, we introduce a transferable region attack (T-RA) by adapting spectrum transformation (ST)~\cite{long2022frequency}. While spectrum transformation was initially devised to improve adversarial attacks targeting {\em image classifiers}, we have discovered that its effectiveness extends to attacking SAM variants as well.
% \gang{The intuition is not clearly explained.}
% \gang{move 3.3 here, explain what is new.
% This idea has been applied in X. Here, we adapt it to do Y. A few sentences on the extra implementation details. 
% }\Zhengyuan{revised}

% \gang{moved to old 3.3 here}
% \paragraph{Spectrum Transformation (ST)}
% \Zhengyuan{revised}

\begin{algorithm}[t]
\caption{Transferable Region Attack (T-RA)}\label{alg:revised_pgd_ssa}
\textbf{Input:} SAM model $f$, image $\boldsymbol{x}$, sampled points $P$ in the region, original segmentation $y$, perturbation limit $\epsilon$, negative threshold $neg\_th$, number of steps $N$, number of spectrum transformed samples $M$, PGD attack step size $\alpha$, spectrum transformation hyperparameters $\rho$ and $\eta$ \\
\textbf{Output:} \text{Adversarial image} $\boldsymbol{x}^\prime$
\begin{algorithmic}[1]
\Procedure{T-RA}{}
    \State $ \mathcal{L}_{\text{best}} \gets \infty $
    \State $\boldsymbol{x}^\prime \gets \boldsymbol{x}$
    \State $\delta \gets 0$
    \For{$step = 1$ To $N$}
        \State $\delta_{sum} \gets 0$
        \For{$i = 1$ To $M$}
            \State $\boldsymbol{x}_{1} \gets \text{ST}(x, \rho, \eta) + \delta$

            \State $L \gets L_{SRA}(x_1, P)$
            \State $\delta_{temp} \gets \text{sign}({\frac{\partial L}{\partial x_1}})*\alpha$
            \State $\delta_{sum} \gets \delta_{sum} + \delta_{temp} $ 
        \EndFor
        \State $\delta \gets \delta_{sum} / M + \delta$
        \State $x' \gets  \text{Clip}(x+\delta, x-\epsilon, x+\epsilon)$
    \EndFor
    \State \textbf{return} $\boldsymbol{x}^\prime$
\EndProcedure
\end{algorithmic}
\end{algorithm}

To improve transferability, model augmentation~\cite{lin2019nesterov} utilizes loss-preserving transformations on the image to avoid the adversarial attack overfitting the current model. Spectrum Transformation (ST)~\cite{long2022frequency} is a form of model augmentation that perturbs the image in the frequency domain. 
 % \gang{is DCT about adding perturbation or simply transferring an image to the frequency domain?}
The intuition is the following: regions of high and low frequency in the image correspond to areas of significant and minor pixel variations respectively. High-frequency areas often represent edges and textures, indicative of rapid pixel intensity changes, while low-frequency areas denote smoother, homogeneous regions that often encompass entire objects. Different SAM variants depend on different frequency domains of interest to make predictions.
By manipulating the spectrum of the image, the idea is to simulate and exploit feature variations of different victim models to enhance the transferability of adversarial attacks. The Discrete Cosine Transform (DCT) and inverse Discrete Cosine Transform (iDCT) are utilized to transform the image back and forth in the spatial and spectrum space. The transformation is formalized as follows:
\begin{equation}
    ST(x, \rho, \eta) = \textit{iDCT}(\textit{DCT}(x + \eta) \odot M(\rho)),
\end{equation}
where $x$ denotes the original image, $\eta$ is a noise vector drawn from a normal distribution $\mathcal{N}(0, \sigma^2 \mathbf{I})$, and $M(\rho)$ a mask with elements sampled from a uniform distribution $\mathcal{U}(1-\rho, 1+\rho)$. The operation $\odot$ represents element-wise multiplication. Note that $\rho$ controls the strength of perturbation. When $\rho$ is too high, the resulting image may not preserve the semantics of the original image; in contrast, when $\rho$ is too low, the adversarial example loses transferability.
% \Zhengyuan{revised}

% \gang{What is iDCT and what is ST?}
% \gang{explain $\rho$
% high-$\rho$: means perturb
% Intuition: how does ST make the attack more transferable? 
% ???
% how does $\rho$ control this process? trade-off of overfitting to a whitebox model and effective noise.
% high $\rho$ means? perturbation less overfitted? 
% }

We detail the implementation of our T-RA attack in \cref{alg:revised_pgd_ssa}. The algorithm takes in a set of parameters including the target model $f$, an original image $\boldsymbol{x}$, and a pre-defined attack region $R$. The process iterates over a predefined number of steps ($N$) and a predefined number of spectrum-transformed samples ($M$), dynamically adjusting the perturbation $\delta$ to minimize the loss $L$, thereby maximizing the adversarial effect. The parameter $\epsilon$ represents the maximum allowable change for each pixel value in the image, ensuring that the perturbations remain imperceptible to the human eye. The algorithm applies spectrum simulations, transforming the image $\boldsymbol{x}$ into $\boldsymbol{x}_1$ (line 8), for improving the transferability of the final adversarial example $\boldsymbol{x}'$. Following ~\cite{zhang2023attack}, we use PGD for attack optimization.

%% file: sec/4_experiments.tex
\section{Evaluation}
\label{sec:exp}
% In this section, we evaluate the robustness of the SAM model against adversarial attacks. 

\subsection{Experimental Setup}
Our experiments are conducted primarily using two variants of the SAM models~\cite{kirillov2023segment}: ViT-B (91M parameters), and ViT-H (636M parameters). A third ViT-L model (308M parameters) will be used only for selective experiments. More specifically, ViT-B will be used for white-box evaluation. Then for the the black-box evaluation, we run transferred attacks from the smaller ViT-B model to the larger ViT-H model. Finally, in \cref{sec:cross}, we further explore the attack transferability to four more SAM variants. 

We evaluate both the sample-based region attack (S-RA) and practical region attack (T-RA). We begin by defining a target region within the image and then test {\em clean images} with both ViT-B and ViT-H models. We set the width and height of regions to be one-third of the original image. On the clean image, we perform click-based segmentation on a randomly selected point ($p$) within the predefined region, resulting in a segmentation mask termed $Mask_{clean}$. Subsequently, we apply the attack method (either S-RA or T-RA) to attack the images and retest them at the same point ($p$) in both ViT-B and ViT-H models. This attacked image yields another segmentation mask $Mask_{adv}$. A comparative analysis of $Mask_{clean}$ and $Mask_{adv}$ is conducted to assess the efficacy of the attack.
% By incorporating the T-RA methodology, we further enhance our attack strategy. This advanced approach involves executing the  T-RA on the ViT-H model, specifically focusing on the same region and point used in the previous tests. The objective is to evaluate the impact of combined attack methods under black-box conditions.

\paragraph{Baseline.}
Regarding the baseline, given our attack has a novel threat model, we {\em adapt} Attack-SAM~\cite{zhang2023attack} to our threat model. More specifically, we run Attack-SAM to attack the {\em center of the region}, and then during testing time, the attack is evaluated on a randomly sampled point in the region (following the same region-level attack protocol). We acknowledge that Attack-SAM is not designed for region-level attack---the purpose of the experiment is to show whether the attack optimized for a pre-defined prompt can generalize to other (nearby) points in the region. 

\paragraph{Dataset.}
For our evaluation dataset, we randomly select 200 images from the SA-1B dataset~\cite{kirillov2023segment}. To induce perturbations in the images, we constrain the magnitude of adversarial perturbation by setting the value of $\epsilon$ to four distinct levels: 2/255, 4/255, 8/255, and 16/255. These values serve as upper bounds for the perturbation, ensuring controlled and quantifiable levels of adversarial noise. The experiments are conducted on NVIDIA A100 GPU to systematically assess the robustness of the SAM model against varying degrees of constrained perturbations.

\begin{figure}[t]
\centering
\includegraphics[width=\columnwidth]{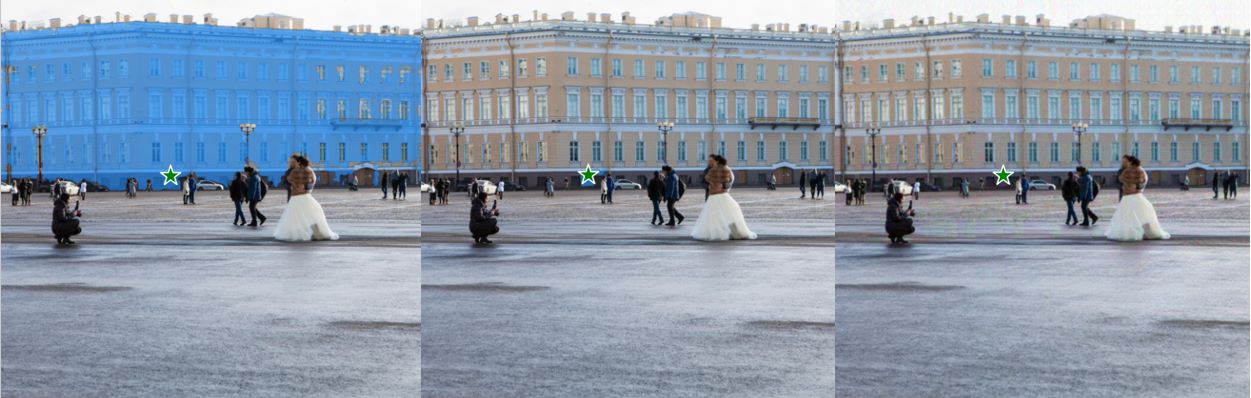}
\caption{Image segmentation results under different attack methods on the ViT-B model. The left image is the original clean image. The middle image is attacked by S-RA and the right image is attached by T-RA. The attack strength is $\epsilon=8/255$.}
\label{fig:white}
% \vspace{-0.1in}
\end{figure}

\begin{table}[t]
\centering
\begin{tabular}{ccccc}
\hline
{$\epsilon$}  & 2/255 & 4/255 & 8/255 & 16/255 \\ \hline
AttackSAM~\cite{zhang2023attack} & 20.56 & 10.48  & 4.28  & 3.69   \\ \hline
   
S-RA & \textbf{2.99}  & \textbf{1.75}  & \textbf{1.52}  & \textbf{1.27}\\ \hline
\end{tabular}
\caption{mIoU (\%) of the white-box experiment on the ViT-B model under the S-RA and AttackSAM~\cite{zhang2023attack} with varying attack strengths ($\epsilon$). 
% A subtle perturbation ($\epsilon=2/255$) from S-RA can already effectively remove most of the mask. 
% We show S-RA outperforms the baseline AttackSAM across all attack strengths.
}
\label{tab:whitemainResult}
\vspace{-0.12in}
\end{table}

\paragraph{Evaluation Metrics.}
Following~\cite{zhang2023attack, chen2017rethinking}, we use the mean Intersection over Union (mIoU) as our primary evaluation metric. mloU is a commonly used metric for evaluating image segmentation. It measures the overlap between the predicted and the ground-truth segmentation masks. Then it takes the mean of the loU values across all test samples. The loU for a single sample is the ratio of the intersection of the predicted segmentation mask $ Mask_{adv}$ and the ground truth mask $Mask_{clean}$ to their union. Mathematically, the mloU is expressed as:
\begin{align}
\text{mIoU}=\frac{1}{N} \sum_{i=1}^N \text{IoU}\left(\text{Mask}_{a d v}^{(i)}, \text { Mask }_{\text {clean }}^{(i)}\right)
\end{align}
where $N$ is the total number of samples in the test set. The value of mIoU ranges from 0 to 1, with a lower value indicating a more effective attack.

\subsection{Qualitative and Quantitative Results}

\begin{figure*}[t]
\centering
\includegraphics[width=\textwidth]{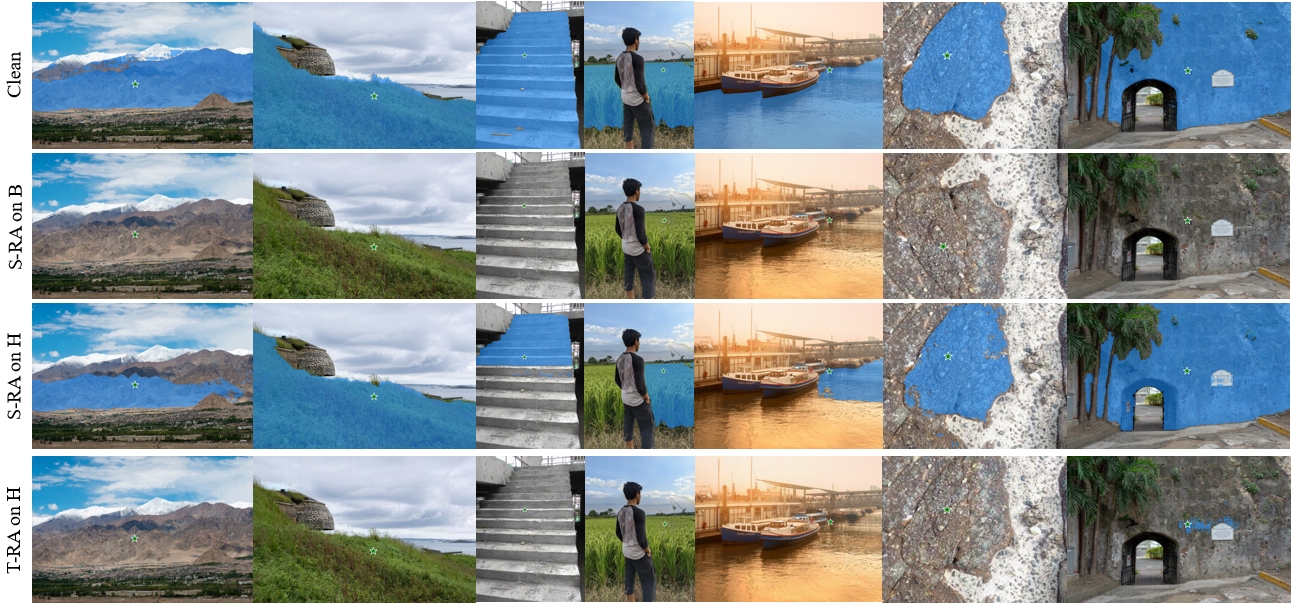}
\caption{Visualization of white-box and black-box attack results (attack strength $\epsilon=$8/255).
% All attacks are optimized/trained on the ViT-B model. 
%
The first row shows the original clean images segmented using the ViT-B model. 
The second row shows S-RA attack (white-box) trained on ViT-B model and the segmentation results on the same ViT-B model. The result confirms the effectiveness of S-RA attack under a white-box setting.  
The third row shows S-RA attack (black-box) trained on ViT-B model and the segmentation results on a different ViT-H model. The result shows the lack of transferability of S-RA under a black-box setting. 
The fourth row shows T-RA attack (black-box) trained on  ViT-B model and the segmentation results on a different ViT-H model. The result shows T-RA transfer well and ViT-H cannot segment correctly under this attack. }
\vspace{-0.1in}
\label{fig:black}
\end{figure*}

We conduct the evaluation under white-box and black-box settings, respectively. 
For the white-box setting, we run the sample-based region attack (S-RA) on the ViT-B model, allowing full access to the model's architecture and parameters (in comparison with the baseline AttackSAM). We confirm the attack is highly successful. 
As shown in \cref{tab:whitemainResult}, a subtle perturbation ($\epsilon=2/255$) from S-RA can already effectively remove most of the mask, hiding the target object from SAM. This is reflected by the minimal overlap between the generated mask and the ground-truth mask (mIoU=2.99\%). Comparing with the baseline, we show S-RA outperforms the adapted AttackSAM~\cite{zhang2023attack} across all attack strengths. \cref{fig:white} shows example attack images from this experiment.

% Our findings demonstrate that both attacks achieved significant success in misleading the model's interpretation of the input images, as evidenced by the adversarial examples and their corresponding segmentation masks illustrated in \cref{fig:white}. The mIoU and F1 scores of white-box experiments are shown in \cref{tab:whitemainResult}. The results demonstrate a decrease in the segmentation capability of the model as $\epsilon$ increases, indicating the effectiveness of the attack in degrading the model's performance. 

\begin{figure*}[t]
\centering
\includegraphics[width=\textwidth]{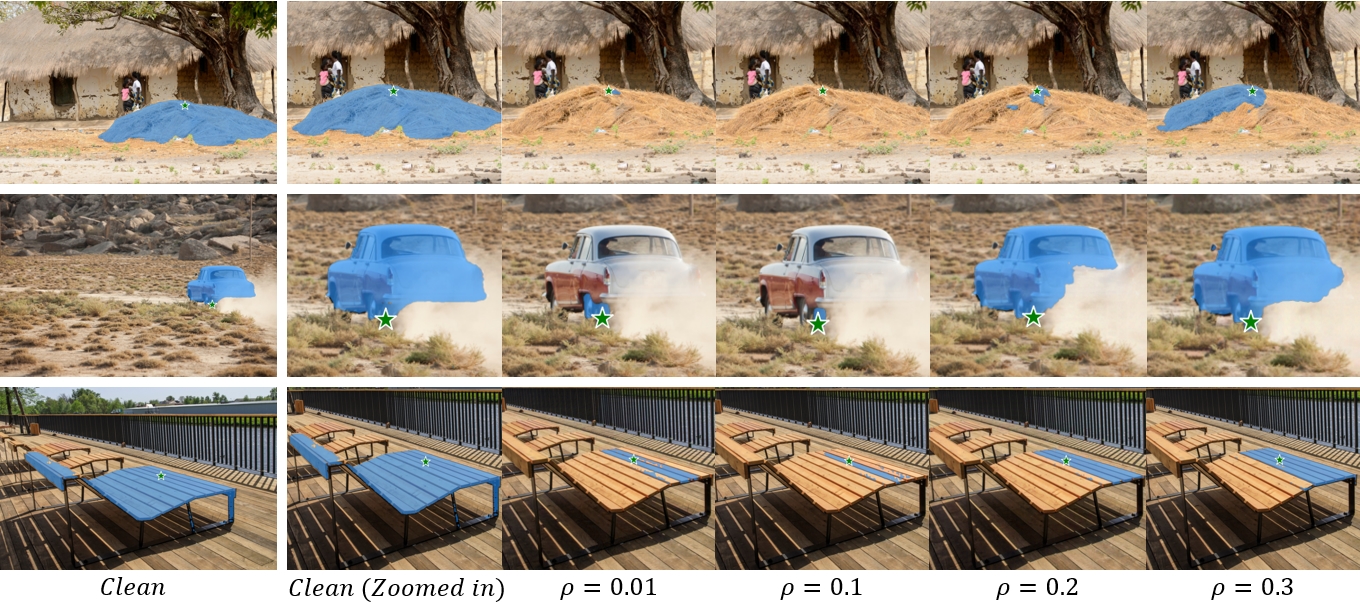}
\caption{Visualization of the segmentation results under different $\rho$ values for the T-RA under a black-box setting (trained ViT-B; tested on the ViT-H). The attack strength is fixed as $\epsilon = 4/255$. 
The first column shows the original clean image's segmentation result. 
The second column is a zoomed-in view to highlight the mask on the clean image. 
The subsequent columns display the segmentation results for $\rho$ values of 0.01, 0.1, 0.2, and 0.3, respectively. We find that $\rho = 0.1$ resulting in the most effective degradation of segmentation accuracy.}
\label{fig:ablation_study}
\vspace{-0.1in}
\end{figure*}

Next, we evaluate the attacks under the black-box setting: the adversarial examples are first computed with the ViT-B model and then used to attack a more complicated ViT-H model. 
The results are shown in \cref{tab:mainResult} and example images are shown in \cref{fig:black}. The result first confirms the lack of transferability of the basic S-RA for black-box attacks (with high mIoU ranging from 31.64\% to 46.32\%). Similarly, the baseline AttackSam~\cite{zhang2023attack} also does not transfer well. Then we show the improved transferability of the T-RA strategy with much lower mIoU. For example, when attack strength is $\epsilon=$8/255, mIoU is below 10\%. The result confirms the effectiveness of T-RA for black-box attacks. 
% \gang{AttackSAM baseline?}

% ViT-H model in a black-box scenario, where the attacker's knowledge of the model's internals is restricted, the effectiveness of the sample-based region attack was notably diminished. The segmentation results from this attack closely resembled those derived from clean images, indicating a limited impact. However, a marked improvement in attack efficacy was observed with the T-RA on the ViT-H model, suggesting its superior capability in circumventing the model's defenses under limited information conditions. The results of the T-RA on the ViT-H model, including the adversarial examples and their segmentation masks, are depicted in \cref{fig:black}. The mIoU and F1 scores of all of our main experiments are shown in \cref{tab:mainResult}.

\begin{table}[t]
\centering
\begin{tabular}{ccccc}
\hline
 $\epsilon$  & 2/255          & 4/255          & 8/255          & 16/255         \\ 
 \hline
{AttackSAM~\cite{zhang2023attack}}  & 59.58          & 48.49          & 43.95          & 32.47   \\  \hline                 
{S-RA}  & 46.32          & 43.32          & 40.75          & 31.64          \\  \hline
T-RA ($\rho$={0.1}) & \textbf{45.01} & \textbf{17.70} & \textbf{9.34}  & \textbf{11.43} \\ 
T-RA ($\rho$={0.3}) & 54.91          & 31.15          & 9.84           & 10.16          \\ \hline
\end{tabular}
\vspace{-0.05in}
\caption{mIoU (\%) result of the AttackSAM~\cite{zhang2023attack}, S-RA and T-RA on under black-box settings (trained on ViT-B, tested on ViT-H). For S-RA, the attack is not as strong as the white-box attack (see \cref{tab:whitemainResult}). Comparing S-RA and T-RA under the black-box setting, T-RA has a stronger attack result with an overall lower mIoU under various settings (especially when $\rho=0.1$). This confirms the effect of spectrum transformation of T-RA.}
\vspace{-0.1in}
\label{tab:mainResult}
\end{table}

\begin{table}
\centering
\begin{tabular}{ccccccc}
\hline
$\rho$  &0.01 &  0.05 & 0.1 & 0.2 & 0.3 \\ \hline
mIoU (\%) &19.32 & 18.31 &  \textbf{17.70}   & 22.98   & 30.51   \\ \hline
\end{tabular}
\vspace{-0.05in}
\caption{mIoU of varying the $\rho$ parameter in the T-RA under the black-box setting (trained on ViT-B; tested on ViT-H), with a fixed attack strength $\epsilon = 4/255$. The results indicate that a $\rho$ value of 0.1 yields the most effective attack, achieving the lowest mIoU.}
\vspace{-0.1in}
% \vspace{-0.1in}
\label{tab:ablation_study}
\end{table}

\subsection{Ablation Study}
\label{sec:ablation}

\paragraph{$\boldsymbol{\rho}$ of the T-RA.} We conduct an ablation study on the \(\rho\) parameter of the T-RA, assessing its impact on the effectiveness of adversarial attacks (trained on ViT-B; tested on ViT-H). The study systematically varies \(\rho\) across a set of values: 0.01, 0.05, 0.1, 0.2, and 0.3, while maintaining a constant \(\epsilon\) value of 4/255.  This investigation allows us to discern the optimal range for \(\rho\). 
% that maximizes the adversarial impact on the model's segmentation output without compromising the imperceptibility of the perturbations to human observers. 
The mIoU percentage results for different \(\rho\) values in the T-RA are presented in \cref{tab:ablation_study}. Additionally, \cref{fig:ablation_study} illustrates the visual differences in the adversarial examples generated with varying \(\rho\) values. The result shows that the attack generally works well under these \(\rho\) values, and the best-performing value is 0.1.

\paragraph{T-RA on ViT-L and ViT-H.} We then extend our investigation of the T-RA to explore its transferability across other ViT models from SAM~\cite{kirillov2023segment}, namely the 308M parameters ViT-L and the 636M parameters ViT-H. Given the substantial variance in model size and complexity, from the 91M parameters ViT-B to these larger architectures, it is imperative to assess the robustness of our adversarial strategy. 
We conduct experiments with \(\epsilon\) values of 8/255 and 16/255, and \(\rho\) parameters set to 0.1 and 0.3. We show the results in \cref{tab:viT_ablation_study}, with the impact of \(\epsilon\) and \(\rho\) settings on the mIoU across the ViT-L and ViT-H models. The results confirm the transferability of both models. Also, the transferability is consistently higher with larger  perturbations (\(\epsilon\) = 8/255).

\begin{table}[t]
\centering
\begin{tabular}{cccccc}
\hline
\multicolumn{2}{c}{$\rho$, $\epsilon$ (/255)}  & 0.1, 4 & 0.1, 8 & 0.3, 4 & 0.3, 8 \\ \hline
{ViT-L} & mIoU (\%) & 13.37  & \textbf{4.40}   & 26.80  & 6.34   \\ \hline
{ViT-H} & mIoU (\%) & 17.70  & \textbf{9.33}   & 30.51  & 9.84   \\ \hline
\end{tabular}
\caption{mIoU (\%) result of the effectiveness T-RA under a black-box setting under various $\epsilon$ and $\rho$ settings. The attack is trained on ViT-B and tested on ViT-L and ViT-H, respectively. 
Bold values highlight the most successful attack configurations. Note that the $\epsilon$ values are represented as ``4'' and ``8'' for brevity, but they correspond to $4/255$ and $8/255$, respectively.}
\label{tab:viT_ablation_study}
\vspace{-0.1in}
\end{table}

\paragraph{Density of Attack Points.} Recall that our attack samples attack points from the target region to compute adversarial examples. Here, we focus on T-RA and investigate the impact of the attack point density on the attack effectiveness. We define the density using a parameter $\lambda$, which represents the number of pixels between consecutive attack points sampled in both horizontal and vertical directions. Recall that we sample $m \times n$ attack points using a grid.  Formally, $m = \frac{W}{\lambda}$ and $n = \frac{H}{\lambda}$.
In our experiments, we test different values of $\lambda$: 50, 60, 70, and 80 pixels. This means that for a given $\lambda$, an attack point is placed every $\lambda$ pixels along both axes, forming a grid-like pattern of attack points within the region. The experiments are carried out using images with $\epsilon = 8/255$ and $\rho = 0.1$ to assess the impact of $\lambda$ on the mIoU metric. The results are presented in \cref{tab:density_study}. As expected, a higher density of sampled points (i.e., lower $\lambda$) leads to a more effective attack. 

\begin{table}
\centering
\begin{tabular}{cccccc}
\hline
$\lambda$  &50 & 60 & 70 & 80  \\ \hline
mIoU (\%) &8.97 & 10.65 & 11.50   & 12.69      \\ \hline
\end{tabular}
\caption{Impact of varying attack point density on the mIoU metric for the T-RA attack. A lower $\lambda$ represents a higher point density.}
\vspace{-0.1in}
\label{tab:density_study}
\end{table}

%% file: sec/5_transfer.tex
\section{Cross Model Transferability}
\label{sec:cross}

In this section, we further evaluate the transferability of our adversarial examples with a broader set of SAM variants with different architectures, under the black-box setting.  
The adversarial examples are generated with $\epsilon = 8/255$ and $\rho = 0.1$ using T-RA, all trained on ViT-B. The testing SAM models include EfficientSAM (S and Ti)~\cite{xiong2023efficientsam}, Fast-SAM (S and X)\cite{zhao2023fast}, MobileSAM~\cite{mobile_sam}, and HQ-SAM (B, L, and H)~\cite{sam_hq}. 
These testing models vary in architecture and complexity, from lightweight versions like EfficientSAM (Ti) to more robust versions like HQ-SAM (H). 

~\cref{fig:transfer} shows the mean Intersection over Union scores at different attack strengths. The results confirm the overall transferability of our attack on these SAM variants. Notably, HQ-SAM (B) exhibits a major drop in mIoU at $\epsilon = 8/255$, indicating a high level of susceptibility to our attack. In contrast, models like MobileSAM maintain higher mIoU scores, demonstrating some level of resilience to stronger attacks. In most models, the segmentation capability weakens as the attack strength increases. However, for HQ-SAM, the weakest segmentation capability is observed at $\epsilon = 8/255$ in two experimental groups. This is likely due to the architectural similarities between HQ-SAM and the original SAM model (see its pattern in \cref{tab:mainResult}), leading to a similar pattern of vulnerability at this attack strength.

\begin{figure}[t]
\centering
\includegraphics[width=\columnwidth]{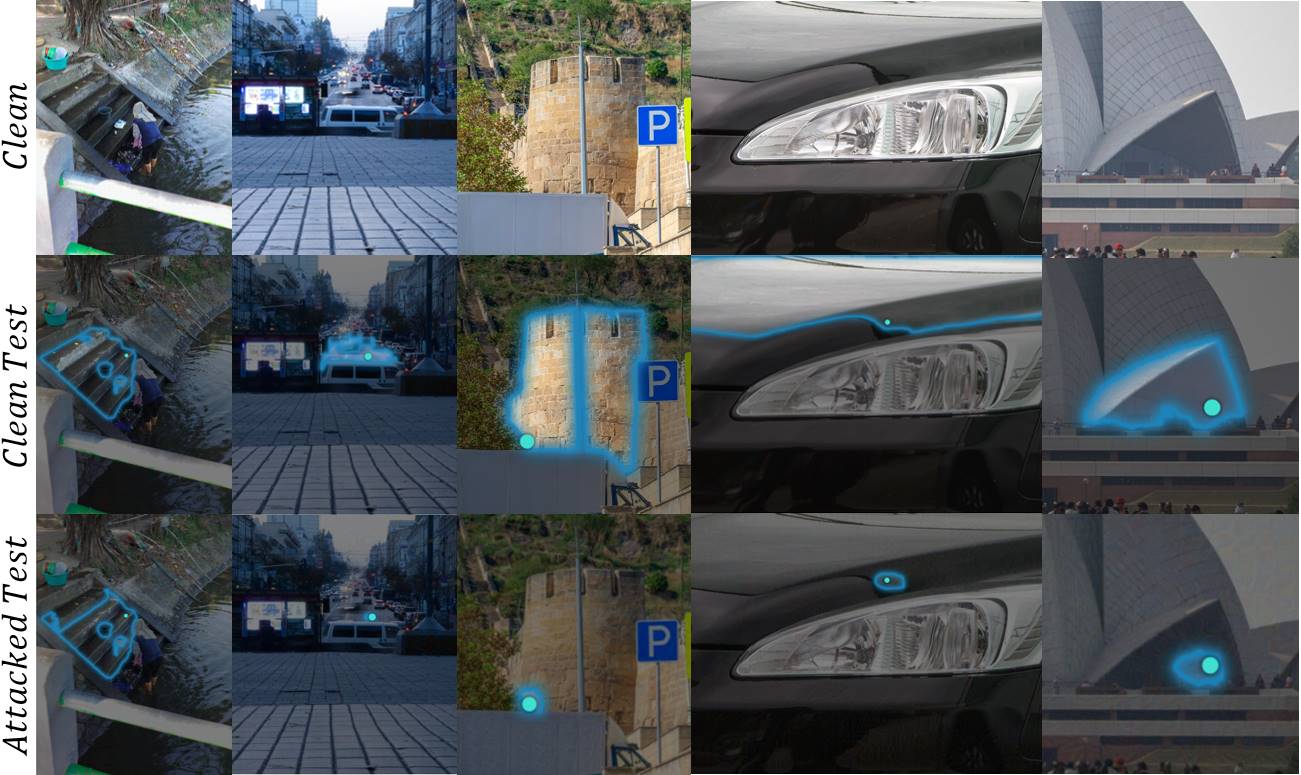}
\caption{Visualization of adversarial examples with $\epsilon = 8/255$ and $\rho = 0.1$, computed on ViT-B and tested on a real-world SAM service. The blue dot is the test point and the highlighted area is the output mask.}
\label{fig:realtest}
\vspace{-0.1in}
\end{figure}

\paragraph{Testing on Real-world SAM Services.} We tested our adversarial example on a real-world SAM Service\footnote{Meta AI SAM: https://segment-anything.com/}, under a black-box setting. Due to limitations in testing through their web interface, a comprehensive quantitative evaluation was not feasible. Instead, we hand-picked a few images, generated adversarial examples with $\epsilon = 8/255$ and $\rho = 0.1$ on the ViT-B model and uploaded the images to the SAM service to manually inspect the result. We observed that online services are indeed more robust (against transferred attacks): they may have implemented countermeasures or image preprocessing steps. For example, not all the points in the attack region can successfully trigger the adversarial effect. We still found some successful images (under the ``point click'' prompt). Examples are shown in ~\cref{fig:realtest}. The experiment was conducted {\em ethically}: these images were uploaded to our personal account and immediately deleted after the tests (i.e., not affecting other users or the service).

% In most models, the segmentation capability weakens as the attack strength increases. However, for HQ-SAM, the weakest segmentation capability is observed at $\epsilon = 8/255$ in two experimental groups. This is likely due to the architectural similarities between HQ-SAM and the original SAM model, leading to a similar pattern of vulnerability at this attack strength.
\begin{table}[t]
\centering
\begin{tabularx}{\columnwidth}{ccccc}
\hline
    $\epsilon$                                        & 2/255 & 4/255 & 8/255 & 16/255 \\ \hline
EfficientSAM (S)~\cite{xiong2023efficientsam}    & 11.80     & 11.50     & 11.62     & \textbf{11.38}      \\ \hline
EfficientSAM (Ti)~\cite{xiong2023efficientsam}    & 8.06     & 7.92     & 7.75     & \textbf{7.13}      \\ \hline
Fast-SAM (S)~\cite{zhao2023fast}                 & 40.06     & 35.14     & 25.74     & \textbf{21.42}      \\ \hline
Fast-SAM (X)~\cite{zhao2023fast}                 & 67.51     & 60.15     & 55.38     & \textbf{37.78}      \\ \hline
MobileSAM~\cite{mobile_sam}                  & 72.92     & \textbf{52.65}     & 57.01     & 58.40      \\ \hline
HQ-SAM (B)~\cite{sam_hq}                         & 51.58     & 21.48     & 2.61     & \textbf{0.01}      \\ \hline
HQ-SAM (L)~\cite{sam_hq}                         & 69.25     & 36.25     & \textbf{22.60}     & 31.47      \\ \hline
HQ-SAM (H)~\cite{sam_hq}                         & 71.01     & 42.11     & \textbf{29.50}     & 37.10      \\ \hline
\end{tabularx}
\caption{mIoU (\%) of transferred attack T-RA with different attack strength $\epsilon$ and a fixed $\rho = 0.1$ on different SAM variants. All attacks are trained by the ViT-B model, and then tested on each of the listed SAM variants under the black-box setting. The lowest mIoU value (i.e., the most successful attack) for each model is highlighted in bold. 
% In most models, the segmentation capability weakens as the attack strength increases. However, for HQ-SAM, the weakest segmentation capability is observed at $\epsilon = 8/255$ in two experimental groups. This is likely due to the architectural similarities between HQ-SAM and the original SAM model, leading to a similar pattern of vulnerability at this attack strength.
}
\label{fig:transfer}
\vspace{-0.1in}
\end{table}

%% file: sec/6_discussion.tex
\section{Conclusion and Future Work}\label{sec:discuss}
In this paper, we introduce a more practical region-level adversarial attack against Segment Anything Models (SAM).
We show that the proposed methods can effectively generate {\em transferable} adversarial examples that compromise SAM's segmentation ability within attacker-defined regions. 
Through extensive experiments, we demonstrate the feasibility of region-level attacks in both white-box and black-box settings and confirm the effectiveness of attacks on multiple SAM variants. The result calls for more robust SAM models to withstand such adversarial threats.

\paragraph{Defense and Future Work.}
Potential defense mechanisms to enhance the robustness of SAM models include applying adversarial training techniques~\cite{madry2017towards,tramer2017ensemble}, the use of input transformation methods to reduce the effectiveness of adversarial perturbations~\cite{guo2017countering,xie2017mitigating,liao2018defense}, and the exploration of novel SAM architectures that are inherently more resistant to adversarial manipulation~\cite{moosavi2019robustness}. Investigating these defenses and their effectiveness in mitigating the threats posed by adversarial attacks on SAM models is an important area for future research.

In this paper, we primarily focus on SAM and its variants that support point prompts. The transferability and effectiveness of our approach on other {\em segmentation models} and other {\em prompt types} remain to be explored, which can be a potential direction for future work. In addition, the real-world impact of our attack may be further influenced by factors such as image quality, image preprocessing, and the presence of countermeasures. 

% Finally, future work may explore ways to train more robust SAM, e.g., using adversarial training techniques or by exploring alternative SAM architectures. 